\documentclass[10pt,twocolumn,letterpaper]{article}

\usepackage[pagenumbers]{cvpr} 

\usepackage{graphicx} 
\usepackage{amsmath}
\usepackage{amssymb}
\usepackage{booktabs}
\usepackage[accsupp]{axessibility}  
\usepackage{colortbl}
\definecolor{mygray3}{gray}{.7}
\definecolor{mygray2}{gray}{.8}
\definecolor{mygray}{gray}{.9}
\usepackage{makecell}
\newcommand{\cmark}{\ding{51}}%
\usepackage{multirow}

\usepackage{graphics}           
\usepackage{times}              
\usepackage{amsmath}            
\usepackage{amssymb}            
\usepackage{graphicx}
\usepackage{algorithm}
\usepackage[noend]{algpseudocode}
\usepackage{booktabs}
\usepackage{color}
\definecolor{instructioncolor}{rgb}{.5,.5,.5}

\usepackage[font=small]{caption}

\def\eqref#1{Eq.~(\ref{#1})}

\makeatletter
\usepackage{xspace}
\DeclareRobustCommand\onedot{\futurelet\@let@token\@onedot}
\def\@onedot{\ifx\@let@token.\else.\null\fi\xspace}

\makeatother

\usepackage{array}
\newcolumntype{L}[1]{>{\raggedright\let\newline\\\arraybackslash\hspace{0pt}}m{#1}}
\newcolumntype{C}[1]{>{\centering\let\newline\\\arraybackslash\hspace{0pt}}m{#1}}
\newcolumntype{R}[1]{>{\raggedleft\let\newline\\\arraybackslash\hspace{0pt}}m{#1}}

\usepackage{colortbl}
\usepackage{makecell}
\usepackage{pifont}%
\usepackage{comment}%

\usepackage[pagebackref,breaklinks,colorlinks]{hyperref}

\usepackage[capitalize]{cleveref}
\crefname{section}{Sec.}{Secs.}
\Crefname{section}{Section}{Sections}
\Crefname{table}{Table}{Tables}
\crefname{table}{Tab.}{Tabs.}

\def\cvprPaperID{1563} 
\def\confName{CVPR}
\def\confYear{2023}

\begin{document}

\title{LiDAR2Map: In Defense of LiDAR-Based Semantic Map Construction Using Online Camera Distillation}

\author{Song Wang, \ \ \ Wentong Li, \ \ \  Wenyu Liu,  \ \ \ Xiaolu Liu, \ \ \ Jianke Zhu$\thanks{Corresponding author is Jianke Zhu.}$\\
	Zhejiang University \\
	{\tt\small \{songw, liwentong, liuwenyu.lwy, xiaoluliu, jkzhu\}@zju.edu.cn}
}
\maketitle

\begin{abstract}
\vspace{-2mm}
Semantic map construction under bird’s-eye view (BEV) plays an essential role in autonomous driving. 
In contrast to camera image, LiDAR provides the accurate 3D observations to project the captured 3D features onto BEV space inherently.
However, the vanilla LiDAR-based BEV feature often contains many indefinite noises, where the spatial features have little texture and semantic cues.
In this paper, we propose an effective LiDAR-based method to build semantic map. Specifically,
we introduce a BEV feature pyramid decoder that learns the robust multi-scale BEV features for semantic map construction, which greatly boosts the accuracy of the LiDAR-based method. To mitigate the defects caused by lacking semantic cues in LiDAR data, we present an online Camera-to-LiDAR distillation scheme to facilitate the semantic learning from image to point cloud.
Our distillation scheme consists of feature-level and logit-level distillation to absorb the semantic information from camera in BEV.
The experimental results on challenging nuScenes dataset demonstrate the efficacy of our proposed LiDAR2Map on semantic map construction, which significantly outperforms the previous LiDAR-based methods over 27.9\% mIoU and even performs better than the state-of-the-art camera-based approaches. Source code is available at: \href{https://github.com/songw-zju/LiDAR2Map}{https://github.com/songw-zju/LiDAR2Map}.
\end{abstract}

\vspace{-2mm}
\section{Introduction}
\label{sec:intro}

\begin{figure}
\centering
\includegraphics[width=0.95 \linewidth]{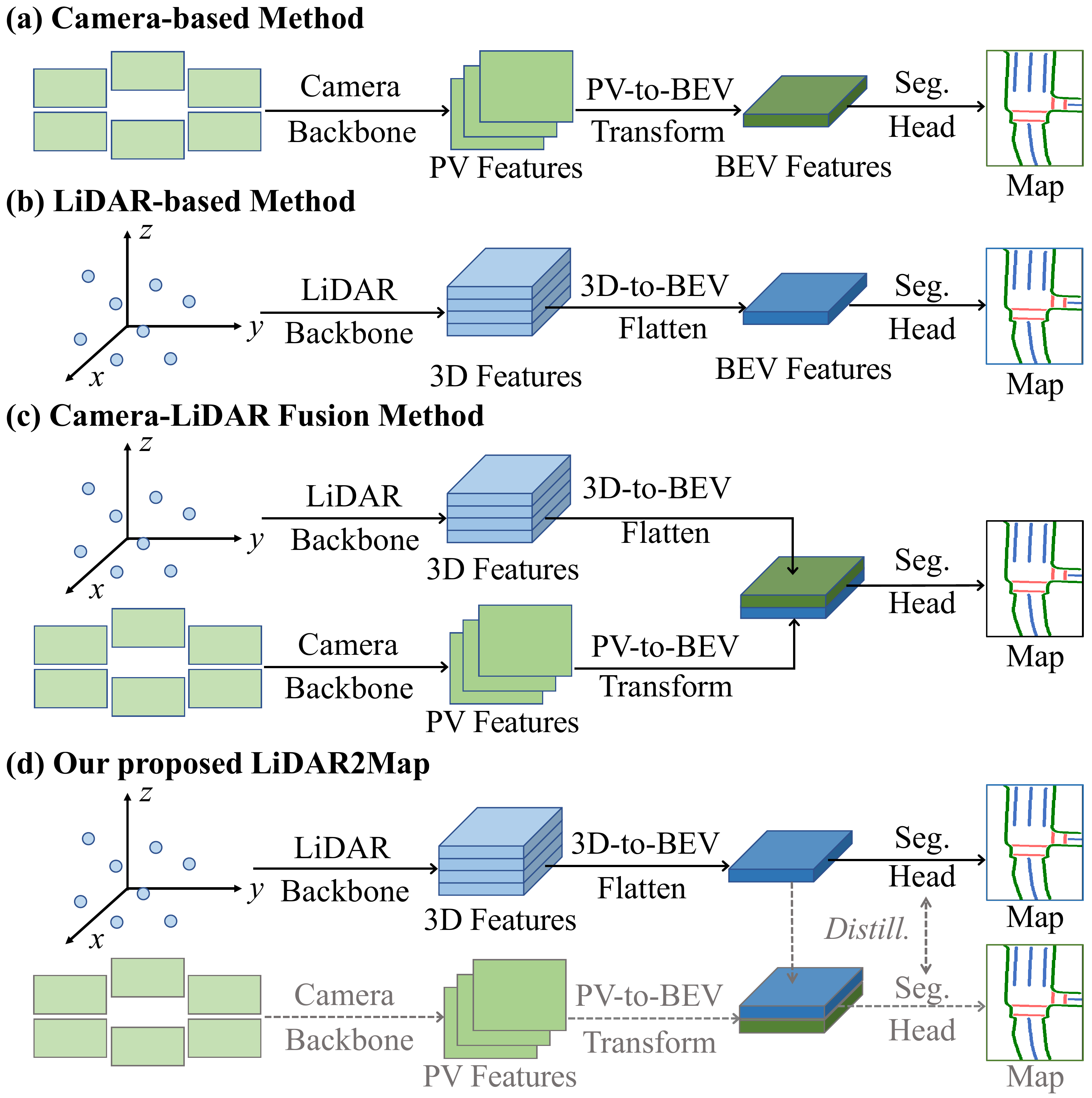}
    \caption{Comparisons on semantic map construction frameworks (camera-based, LiDAR-based, Camera-LiDAR fusion methods) and our proposed LiDAR2Map that presents an effective online Camera-to-LiDAR distillation scheme with a BEV feature pyramid decoder in training.} 
    \label{fig:intro}
    \vspace{-7mm}
\end{figure}

High-definition (HD) map contains the enriched semantic understanding of elements on road, which is a fundamental module for navigation and path planning in autonomous driving. 
Recently, online semantic map construction has attracted increasing attention, which enables to construct HD map at runtime with onboard LiDAR and cameras. 
It provides a compact way to model the environment around the ego vehicle,  which is convenient to obtain the essential information for the downstream tasks.

Most of recent online approaches treat semantic map learning as a segmentation problem in bird’s-eye view (BEV), which assign each map pixel with a category label. As shown in Fig.~\ref{fig:intro}, the existing methods can be roughly divided into three groups, including 
camera-based methods~\cite{peng2023bevsegformer, zhang2022beverse, qin2022unifusion, li2022bevformer, li2022hdmapnet}, 
LiDAR-based methods~\cite{hendy2020fishing, li2022hdmapnet} and Camera-LiDAR fusion methods~\cite{li2022hdmapnet, liu2023bevfusion, salazar2022transfusegrid}.
Among them,   
camera-based methods are able to make full use of multi-view images with the enriched semantic information, which dominate this task with the promising performance.
In contrast to camera image, LiDAR outputs the accurate 3D spatial information that can be used to project the captured features onto the BEV space. By taking advantage of the geometric and spatial information, LiDAR-based methods are widely explored in 3D object detection~\cite{zhou2018voxelnet, lang2019pointpillars, shi2020pv, yin2021center} while it is rarely investigated in semantic map construction. 
HDMapNet-LiDAR~\cite{li2022hdmapnet} intends to directly utilize the LiDAR data for map segmentation, however, it performs inferior to the camera-based models due to the vanilla BEV feature with the indefinite noises. Besides, map segmentation is a semantic-oriented task~\cite{liu2023bevfusion} while the semantic cues in LiDAR are not as rich as those in image. 
In this work, we aim to exploit the LiDAR-based semantic map construction by taking advantage of the global spatial information and auxiliary semantic density from the image features.

In this paper, we introduce an efficient framework for semantic map construction, named LiDAR2Map, which fully exhibits the potentials of LiDAR-based model. Firstly, we present an effective decoder to learn the robust multi-scale BEV feature representations from the accurate spatial point cloud information for semantic map. It provides distinct responses and boosts the accuracy of our baseline model. To make full use of the abundant semantic cues from camera, we then suggest a novel online Camera-to-LiDAR distillation scheme to further promote the LiDAR-based model.
It fully utilizes the semantic features from the image-based network with a position-guided feature fusion module (PGF$^2$M).
Both the feature-level and logit-level distillation are performed in the unified BEV space to facilitate the LiDAR-based network to absorb the semantic representation during the training.  Specially, we suggest to generate the global affinity map with the input low-level and high-level feature guidance for the satisfactory feature-level distillation. 
The inference process of LiDAR2Map is efficient and direct without the computational cost of distillation scheme and auxiliary camera-based branch.
Extensive experiments on the challenging nuScenes benchmark~\cite{caesar2020nuscenes} show that our proposed model significantly outperforms the conventional LiDAR-based method (29.5\% mIoU \textit{vs.} 57.4\% mIoU). It even performs better than the state-of-the-art camera-based methods by a large margin.
 
Our main contributions are summarized as: 1) an efficient framework LiDAR2Map for semantic map construction, where the presented BEV feature pyramid decoder can learn the robust BEV feature representations to boost the baseline of our LiDAR-based model; 2) an effective online Camera-to-LiDAR distillation scheme that performs both feature-level and logit-level distillation during the training to fully absorb the semantic representations from the images; 3) extensive experiments on nuScenes for semantic map construction including map and vehicle segmentation under different settings, shows the promising performance of our proposed LiDAR2Map.

\section{Related Work}
\label{sec:related}
\noindent \textbf{Semantic Map Construction.} 
High-definition (HD) maps have the rich information on road layout, which are essential to autonomous vehicles~\cite{liu2020high, bauer2016using, yang2018hdnet}. Traditional offline approaches to HD map construction require lots of manual annotations and regular updates~\cite{besl1992method, yu2015semantic, zhao2021fidnet, kim2021hd, wang2022meta}, which incur the expensive costs on labeling. Recently, the learning-based methods~\cite{zhou2021automatic, li2022hdmapnet, liu2022vectormapnet} have been proposed to construct semantic map online with camera image and LiDAR point cloud using an end-to-end network, which can be roughly divided into three groups, including camera-based methods, LiDAR-based approaches and Camera-LiDAR fusion methods. Camera-based methods~\cite{peng2023bevsegformer, zhang2022beverse, qin2022unifusion} learn to project the perspective view (PV) features onto BEV space through the geometric prior, which often have the spatial distortions inevitably.
Besides, the camera-based methods rely on high-resolution images and large pre-trained models for better accuracy~\cite{zhang2022beverse, li2022bevformer}, which brings serious challenges to the practical scenarios.
LiDAR-based approaches~\cite{hendy2020fishing, li2022hdmapnet} directly capture the accurate spatial information for the unified BEV feature representation. 
However, they cannot robustly deal with large noises in the vanilla BEV feature.
Camera-LiDAR fusion methods~\cite{liu2023bevfusion, li2022hdmapnet, salazar2022transfusegrid} make use of both the semantic features from camera and geometric information from LiDAR. They achieve better results than those approaches with single modality under the same setting while having the larger computational burden. In this paper, we intend to construct the semantic map from LiDAR point cloud effectively.

\begin{figure*}
\centering
\includegraphics[width=0.9\linewidth]{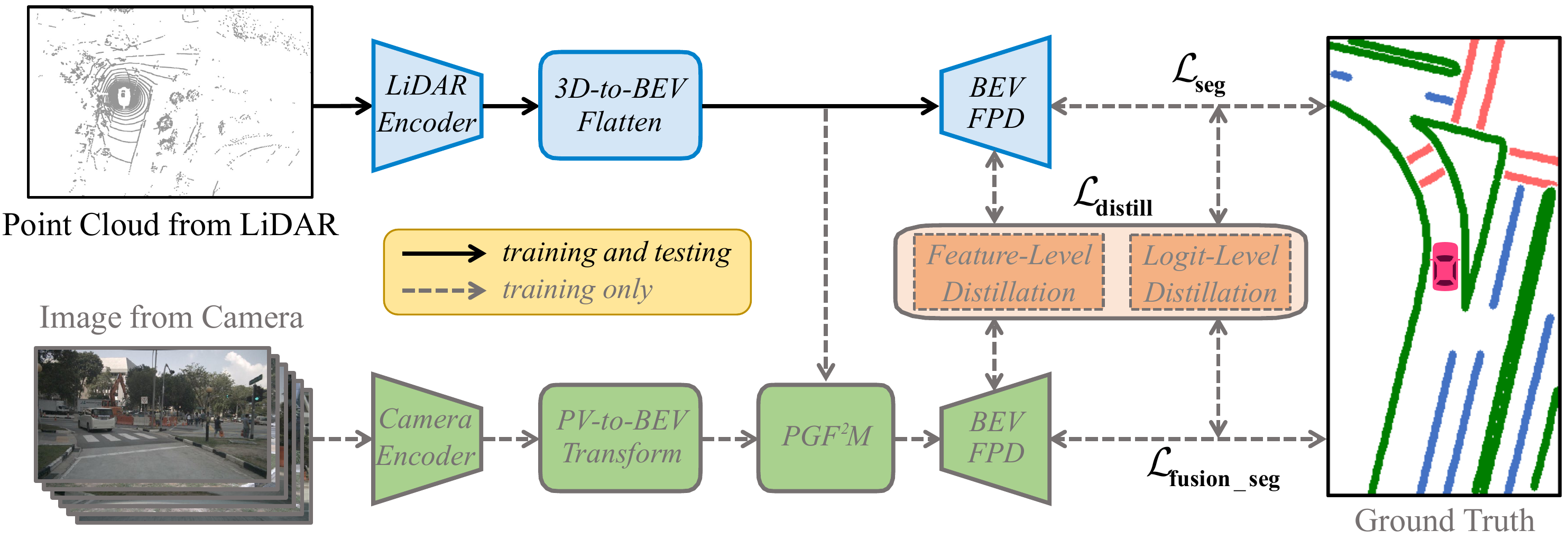}
    \caption{\textbf{Overview of LiDAR2Map Framework.} LiDAR2Map employs the LiDAR-based network as the main branch to encode the point cloud feature with a robust BEV feature pyramid decoder (BEV-FPD) for semantic map construction. During the training, the camera-based branch is adopted to extract the semantic image features. Both feature-level and logit-level distillation are performed to allow LiDAR-branch to benefit from the providing image features without the overhead during inference.}
    \label{fig:framework}
    \vspace{-4mm}
\end{figure*}

\noindent \textbf{Multi-sensor Fusion.} 
Multi-sensor fusion is always a key issue in autonomous driving, among which camera and LiDAR fusion research is the most in-depth. Previous methods obtain the promising performance on 3D detection and segmentation through a point-to-pixel fusion strategy~\cite{zhuang2021perception, vora2020pointpainting, yin2021multimodal}. However, such pipeline requires the correspondences between points and pixels, which cannot fully utilize the information of whole image and all the point cloud. Recently, multi-modal feature fusion in the unified BEV space has attracted some attention~\cite{liu2023bevfusion, liang2022bevfusion}. Converting the semantic features from camera into a BEV representation can be better integrated with spatial features from LiDAR~\cite{philion2020lift, pan2020cross}. This provides the enriched information for downstream tasks like planning and decision-making. However, the fusion of multi-sensor may increase the computational burden on the deployment. In this work, we exploit an effective online Camera-to-LiDAR distillation scheme to fully absorb the semantic features for LiDAR-based branch.

\noindent \textbf{Cross-modal Knowledge Distillation.} 
Knowledge distillation is originally proposed for model compression~\cite{hinton2015distilling}, where knowledge can be transferred from a pre-trained model to an untrained small model. In addition to logit-level distillation~\cite{cho2019efficacy, furlanello2018born, zhao2022decoupled}, feature-level distillation has received more attention~\cite{romero2014fitnets, heo2019comprehensive, heo2019knowledge, yang2022masked}. Cross-modal knowledge distillation has been validated in many tasks such as LiDAR semantic segmentation~\cite{jaritz2020xmuda, yan20222dpass}, monocular 3D object detection~\cite{chong2022monodistill}, 3D hand pose estimation~\cite{yuan20193d} and 3D dense captioning~\cite{yuan2022x}. In this work, we introduce both feature-level and logit-level distillation on BEV representation.

\section{LiDAR2Map}

\subsection{Overview}
In this work, we aim to explore the potentials of an efficient LiDAR-based model for semantic map construction.
Different from the previous LiDAR-based methods~\cite{hendy2020fishing, li2022hdmapnet}, we introduce an effective BEV feature pyramid decoder to learn the robust representations from the spatial information of point cloud. To enhance the semantic information of single LiDAR modality, we take into account of the images through an online distillation scheme on the BEV space that employs the multi-level distillation during the training. 
In the inference stage, we only preserve the LiDAR branch for efficient semantic map prediction.
Fig.~\ref{fig:framework} shows the overview of our proposed LiDAR2Map framework.

\subsection{Map-Oriented Perception Framework}
\noindent \textbf{Multi-Modal Feature Extractors.} 
LiDAR sensor typically outputs a set of unordered points, which cannot be directly processed by 2D convolution. We investigate the most commonly used backbones in 3D object detection, including PointPillars~\cite{lang2019pointpillars} and VoxelNet~\cite{zhou2018voxelnet}, which can extract the effective 3D features $\mathbf{F}_{\text {LiDAR}}^{\text {3D}}$ from LiDAR point cloud.
Specifically, PointPillars converts the raw point cloud into multiple pillars, and then extracts features from pillar-wise point cloud by 2D convolution. VoxelNet directly voxelizes the point cloud first and uses the sparse convolution to build 3D network to encoder the better 3D feature representation. 
Then, the unified BEV representation $\mathbf{F}_{\text {LiDAR}}^{\text {BEV}}$ is obtained by pooling the 3D features $\mathbf{F}_{\text {LiDAR}}^{\text {3D}}$.

Besides, we build another network branch to encode the pixel-level semantic features in perspective view from the images, which is used in our presented online distillation scheme (see Sec.~\ref{bevdis}). 
As in~\cite{philion2020lift}, we adopt a similar 2D-3D transformation manner. 
Firstly,  
we extract the perspective features $\mathbf{F}_{\text {Camera}}^{\text {PV}}$ from each input image $\mathbf{I}  \in  \mathbb{R}^{3 \times H \times W}$ by 2D convolution and predict the depth distribution of $D$ equally spaced discrete points associated with each pixel. 
Secondly, we assign the perspective features $\mathbf{F}_{\text {Camera}}^{\text {PV}}$ to $D$ points along the camera ray direction to obtain a $D \times H \times W$ pseudo point cloud features $\mathbf{F}_{\text {Camera}}^{\text {3D}}$.
Finally, the pseudo point cloud features are flatten to the BEV space $\mathbf{F}_{\text {Camera}}^{\text {BEV}}$ through the pooling as the LiDAR branch.

\noindent \textbf{BEV Feature Pyramid Decoder.} 
BEV features are regarded as the unified representation in our framework, which can absorb both the geometric structure from LiDAR and semantic features from the images. Based on BEV features, the current LiDAR-based method in~\cite{li2022hdmapnet} employs a fully connected layer as the segmentation head to obtain segmentation results directly. Since the vanilla BEV feature from LiDAR backbone contains the ambiguous noise response, it obtains the inferior performance compared with camera-based models~\cite{peng2023bevsegformer, zhang2022beverse}.

\begin{figure}
\centering
\includegraphics[width=0.9 \linewidth]{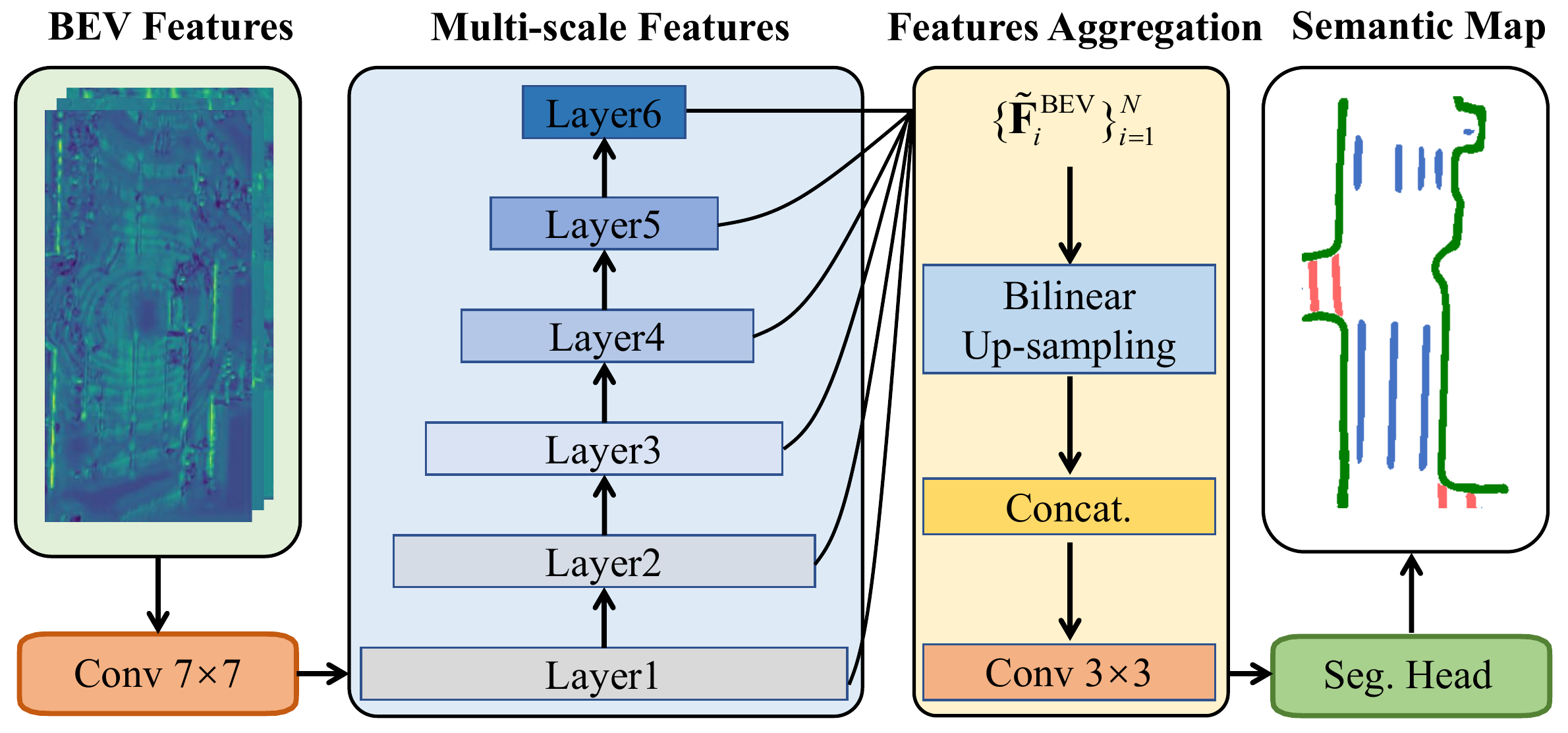}    
\caption{Illustration of \textbf{BEV Feature Pyramid Decoder (BEV-FPD)}. BEV-FPD collects the multi-scale BEV features from six layers to perform the feature aggregation for semantic map with a segmentation head.}
\label{fig:bevdecoder}
    \vspace{-4mm}
\end{figure}

In this work, we develop a BEV feature pyramid decoder (BEV-FPD) to capture the multi-scale BEV features with less noises from LiDAR data for better semantic map construction.  
Fig.~\ref{fig:bevdecoder} shows the architecture of the BEV-FPD. 
Based on the BEV features $\mathbf{F}^{\text {BEV}}$ from the LiDAR or camera branch, we firstly perform $7\times7$ convolution on the BEV features to generate the global features with the large receptive field. The multi-scale BEV features $\{ {\bf{\tilde F}}_i^{{\rm{\text{BEV}}}}\} _{i = 1}^N$ are obtained by the six successive layers, and each layer consists of two standard residual block~\cite{he2016deep} to better transmit the feature representation. The $N$-scale features $\{ {\bf{\tilde F}}_i^{{\rm{\text{BEV}}}}\} _{i = 1}^N$ represent the different level of semantic features in the BEV space. As the feature size decreases, the number of channels increases. The bilinear interpolation is used to up-sample the each low-resolution semantic maps and obtain the feature representations with the same resolution. We then concatenate the feature maps at all scales with the same resolution to perform the multi-scale feature aggregation. The final semantic map is obtained by a segmentation head with the \textit{softmax} function to account for the probability distribution of each category.
As the layer number increases, the corresponding BEV features ${\bf{\tilde F}}_i^{{\rm{\text{BEV}}}}$ can better capture the robust spatial features with accurate responses. It plays an essential role in improving our proposed LiDAR-based model (see Sec.~\ref{ablation_bevdecoder}).

\subsection{Online Camera-to-LiDAR Distillation}
\label{bevdis}
To enhance the semantic representation for our LiDAR-based model, we introduce an effective online Camera-to-LiDAR distillation scheme in BEV space, which enables the LiDAR-based branch to learn the semantic cues from the images. It consists of three components, including Position-Guided Feature Fusion Module (PGF$^2$M), Feature-level Distillation (FD) and Logit-level Distillation (LD).

\begin{figure}
\centering
\includegraphics[width=0.9 \linewidth]{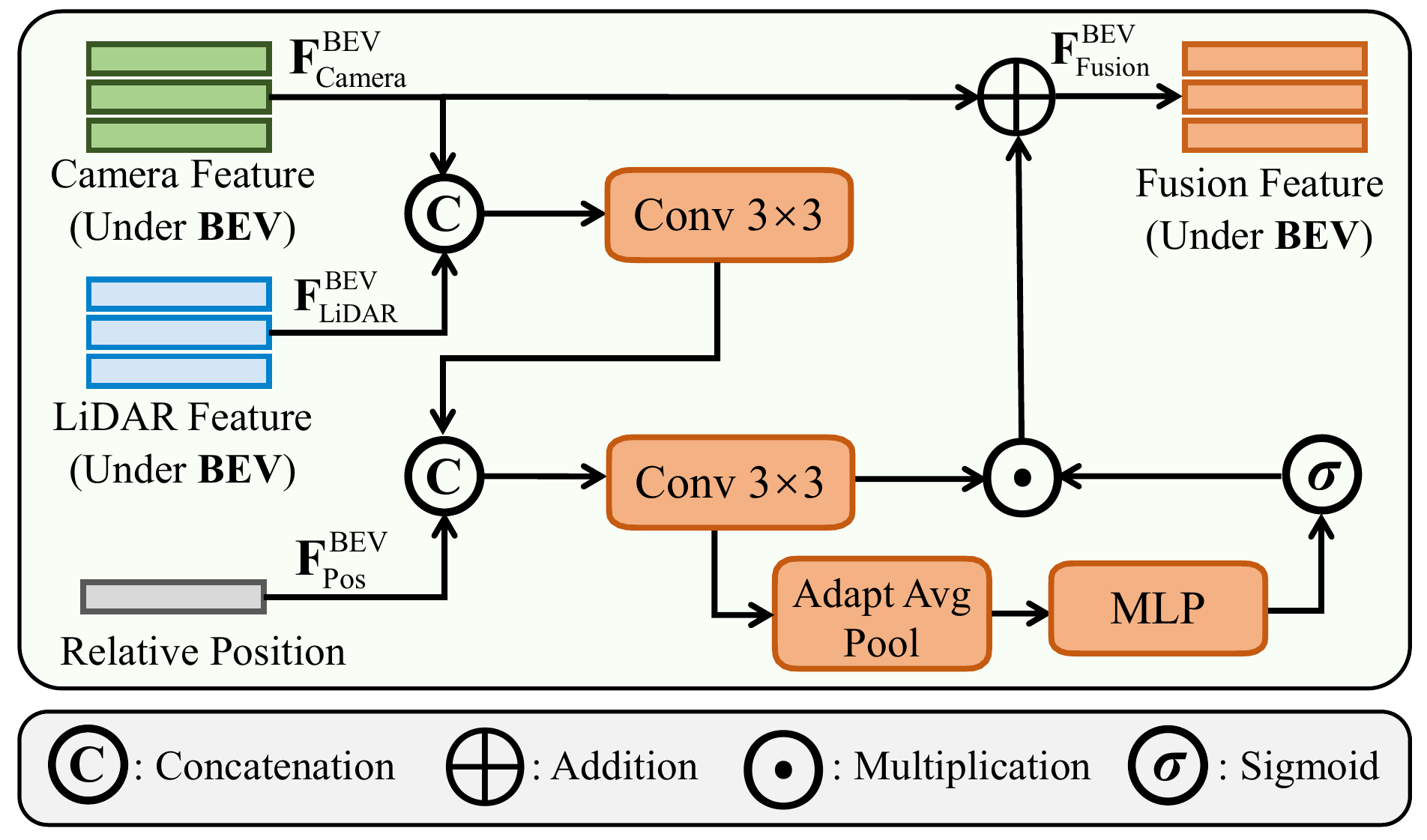}
    \caption{Illustration of \textbf{Position-Guided Feature Fusion Module (PGF$^2$M)}. In PGF$^2$M, the camera image features and LiDAR features in BEV space are in integration with the relative position information. }
    \label{fig:attfusion}
    \vspace{-4mm}
\end{figure}

\noindent \textbf{Position-Guided Feature Fusion Module.} PGF$^2$M  is introduced to better fuse the features from camera and LiDAR in BEV space, as shown in Fig.~\ref{fig:attfusion}. 
Firstly, we concatenate the BEV features along the channel dimension between two modalities, \textit{i.e.} LiDAR point cloud feature $\mathbf{F}_{\text {LiDAR}}^{\text {BEV}}$ and camera image feature $\mathbf{F}_{\text {Camera}}^{\text {BEV}}$. Then, we perform the preliminary fusion through a $3 \times 3$ convolutional layer to obtain $\mathbf{F}_{\text {Fusion\_s1}}^{\text {BEV}}$ as below,
\begin{equation}
\mathbf{F}_{\text {Fusion\_s1}}^{\text {BEV}}=\operatorname{Conv\;3 \times 3}\left(\left[\mathbf{F}_{\text {Camera}}^{\text {BEV}}, \mathbf{F}_{\text {LiDAR}}^{\text {BEV}}\right]\right).
\end{equation}

\noindent
Secondly, we calculate the relative coordinates of $x$-axis and $y$-axis $\mathbf{F}_{\text {Pos}}^{\text {BEV}}$ with the same size. Then, we concatenate it with the fusion result $\mathbf{F}_{\text {Fusion\_s1}}^{\text {BEV}}$ at the previous stage along the channel dimension to encode the spatial information, and perform $3 \times 3$ convolution:
\begin{equation}
\mathbf{F}_{\text {Fusion\_s2}}^{\text {BEV}}=\operatorname{Conv\;3 \times 3}\left(\left[\mathbf{F}_{\text {Fusion\_s1}}^{\text {BEV}}, \mathbf{F}_{\text {Pos}}^{\text {BEV}}\right]\right).
\end{equation}

\noindent
$\mathbf{F}_{\text {Fusion\_s2}}^{\text {BEV}}$ is further fed into an attention layer that is composed of a 2D adaptive average pooling, two-layer MLP and a $sigmoid$ function to build the global pixel affinity. Thus, its result $\mathbf{F}_{\text {Fusion\_s3}}^{\text {BEV}}$ is obtained by
\begin{equation}
\mathbf{F}_{\text {Fusion\_s3}}^{\text {BEV}}=\sigma\left(\operatorname{MLP}\left(\operatorname{Avg}(\mathbf{F}_{\text {Fusion\_s2}}^{\text {BEV}})\right)\right) \odot \mathbf{F}_{\text {Fusion\_s2}}^{\text {BEV}}.
\end{equation}
\noindent
Finally, we add $\mathbf{F}_{\text {Fusion\_s3}}^{\text {BEV}}$ with the original BEV feature from camera $\mathbf{F}_{\text {Camera}}^{\text {BEV}}$ to obtain the fusion features $\mathbf{F}_{\text {Fusion}}^{\text {BEV}}$:
\begin{equation}
\mathbf{F}_{\text {Fusion}}^{\text {BEV}}=\mathbf{F}_{\text {Camera}}^{\text {BEV}}+\mathbf{F}_{\text {Fusion\_s3}}^{\text {BEV}}.
\end{equation}

\noindent \textbf{Feature-level Distillation.}
To facilitate the LiDAR branch to absorb the rich semantic features from the images, we take advantage of the multi-scale BEV features 
$\{ {\bf{\tilde F}}_i^{{\rm{\text{BEV}}}}\} _{i = 1}^N$ from BEV-FPD for the feature-level distillation.
Generally, it is challenging to directly distill high-dimensional features between camera and LiDAR modalities, which lack the global affinity of BEV representation. The straightforward feature distillation on these dense feature often fails to achieve the desired results. 
To address this issue, we employ the tree filter~\cite{song2019learnable, liang2022tree} as the transform function $\mathcal{F}$ to model the long-range dependencies of dense BEV features in each modality by minimal spanning tree.
Specifically, the shallow pillar/voxel features $\mathbf{F}_{\text{low}}^{\text {BEV}}$ from LiDAR backbone and multi-scale BEV features $\{ {\bf{\tilde F}}_i^{{\rm{\text{BEV}}}}\} _{i = 1}^N$  are treated as the low-level and high-level input guidance of tree filter. 
With these low-level and high-level guidance, 
the feature transform is performed by tree filter in the cascade manner to obtain the global affinity map $\mathbf{M}^{\text {BEV}}_i$ for the corresponding $i$-th scale BEV features $ {\bf{\tilde F}}_i^{{\rm{\text{BEV}}}}$ as following,
\begin{equation}
\mathbf{M}_{i}^{\text {BEV}}=\mathcal{F}\left(\mathcal{F}\left({\bf{\tilde F}}_i^{{\rm{\text{BEV}}}}, \mathbf{F}_{\text{low}}^{\text {BEV}}\right), {\bf{\tilde F}}_i^{{\rm{\text{BEV}}}}\right).
\end{equation}
We compute the affinity similarity between each $\mathbf{M}_{\text{LiDAR}, i}^{\text {BEV}}$ from the LiDAR branch and  $\mathbf{M}_{\text{Fusion}, i}^{\text {BEV}}$ of the Camera-LiDAR fusion branch to achieve the feature-level distillation.
More specifically, a simple $L_1$ distance is used to accumulate them at all the scales as below, 
\begin{equation}
\mathcal{L}_{\mathbf{feature}}= \sum_{i=1}^{N} \left\|\mathbf{~M}_{\text{Fusion}, i}^{\text {BEV}}-\mathbf{M}_{\text{LiDAR}, i}^{\text {BEV}}~\right\|_{1}.
\end{equation}
We employ $\mathcal{L}_{\mathbf{feature}}$ as one of the loss terms to enable the LiDAR-based branch to benefit from the image feature implicitly through the network optimization.

\begin{figure}
\centering
\includegraphics[width=0.9 \linewidth]{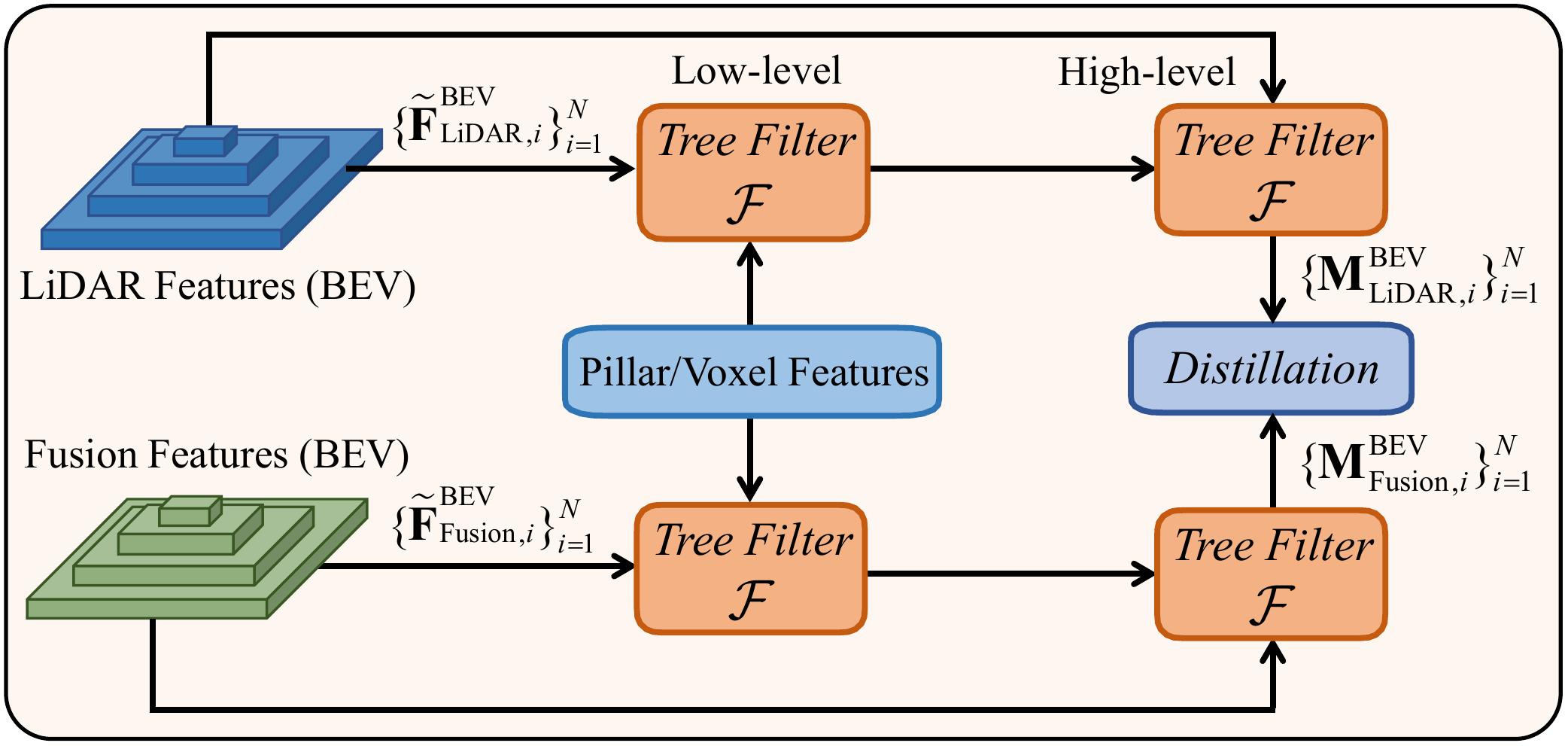}
    \caption{Illustration of \textbf{Feature-level Distillation}. Multi-scale BEV features are fed into two successive tree filters to generates the affinity map with the low-level and high-level guidance. The feature-level distillation is performed on generated affinity maps between LiDAR and fusion branch.}
    \label{fig:fld}
    \vspace{-4mm}
\end{figure}

\noindent \textbf{Logit-level Distillation.}
The semantic map predictions of segmentation head represent the probability distribution of each modality. We further suggest the logit-level distillation to make the LiDAR-based ``Student'' prediction learn from the soft labels generated by Camera-LiDAR fusion model as a ``Teacher''.

Through BEV-FPD with the segmentation head, the corresponding semantic map predictions $\mathbf{P}_{\text {LiDAR}}^{\text {BEV}}$ and $\mathbf{P}_{\text {Fusion}}^{\text {BEV}}$  can be obtained.
As in~\cite{jaritz2020xmuda}, we adopt KL divergence to measure the similarity on the probability distribution, which makes the $\mathbf{P}_{\text {LiDAR}}^{\text {BEV}}$ of LiDAR closer to  $\mathbf{P}_{\text {Fusion}}^{\text {BEV}}$ of fusion  ``Teacher'' as below,
\begin{equation}
\begin{aligned}
\mathcal{L}_{\mathbf{logit}} &=\boldsymbol{D}_{\mathrm{KL}}\left(\mathbf{P}_{\text {Fusion}}^{\text {BEV}} \| \mathbf{P}_{\text {LiDAR}}^{\text {BEV}}\right). 
\label{eq:kld}
\end{aligned}
\end{equation}

\subsection{Training and Inference}
\noindent \textbf{Overall Loss Function for Training.}
In this work, we treat the semantic map construction task as a pixel-level classification problem with segmentation loss in network optimization. Overall, the total training loss of our proposed framework consists of three terms:
\begin{align}
    {\mathcal{L}}={\mathcal{L}}_{\mathbf{seg}}+{\mathcal{L}}_{\mathbf{fusion\_seg}}+ {\mathcal{L}}_{\mathbf{distill}},
\end{align}
where ${\mathcal{L}}_{\mathbf{seg}}$ and ${\mathcal{L}}_{\mathbf{fusion\_seg}}$ are the segmentation losses of LiDAR-branch and Camera-LiDAR fusion branch, respectively. 
$\mathcal{L}_{\mathbf{distill}}$ consists of $\mathcal{L}_{\mathbf{feature}}$ and $\mathcal{L}_{\mathbf{logit}}$ for online Camera-to-LiDAR distillation.

The segmentation loss for semantic map construction is composed of two items including ${\mathcal{L}}_{\mathbf{ce}}$ and ${\mathcal{L}}_{\mathbf{ls}}$ as following,
\begin{align}
    {\mathcal{L}}_{\mathbf{seg}}={\mathcal{L}}_{\mathbf{ce}}+{\mathcal{L}}_{\mathbf{ls}},
\end{align}
where ${\mathcal{L}}_{\mathbf{ce}}$ is the cross-entropy loss. ${\mathcal{L}}_{\mathbf{ls}}$ is employed to maximize the Intersection-over-Union (IoU) score as below,
\begin{align}
    \mathcal{L}_{\mathbf{ls}}=\frac{1}{|C|} \sum_{c \in C} \overline{\Delta_{J_{c}}}(\mathbf{m}(c)),
\end{align}
where $|C|$ is the total number of classes. $\mathbf{m}(c)$ denotes the vector of pixel errors on class $c \in C$. $\overline{\Delta_{J_{c}}}$ is the Lov\'asz extension \cite{berman2018lovasz} for $\mathbf{m}(c)$ as the surrogate loss. The calculation of ${\mathcal{L}}_{\mathbf{fusion\_seg}}$ is the same as ${\mathcal{L}}_{\mathbf{seg}}$.

\noindent \textbf{Inference.}
The LiDAR-based branch is fully optimized during training, which not only captures the spatial geometric features but also absorbs the enriched semantic information from the camera images.
It is worthy of noting that we only preserve the LiDAR-branch for the predictions. The inference process is direct and efficient without incurring the computational cost on distillation and the camera-based branch.

\section{Experiments}

\begin{table*}
\begin{center}
    \caption{Performance comparison on the validation set of nuScenes with the $60\text{m} \times 30\text{m}$ setting for map segmentation. ``$*$'' means the results reported from HDMapNet~\cite{li2022hdmapnet}. ``${\dag}$'' denotes the results reported from UniFusion~\cite{qin2022unifusion}. } 
    \label{tab:mapseg}
    \scalebox{0.82}{
    \begin{tabular}{lccccccc}
    \toprule
    {Method} & {Image Size} & {Modality} & {Backbone}  & Divider & Ped Crossing & Boundary & mIoU  \\ \midrule
    VPN$^{*}$~\cite{pan2020cross} & 352$\times$128 & Camera & EfficientNet-B0~\cite{tan2019efficientnet} & 36.5  & 15.8 & 35.6 & 29.3 \\
    Lift-Splat$^{*}$~\cite{philion2020lift} & 352$\times$128 & Camera & EfficientNet-B0 & 38.3 & 14.9   & 39.3 & 30.8  \\
    HDMapNet-Camera~\cite{li2022hdmapnet} & 352$\times$128 & Camera & EfficientNet-B0  & 40.6 & 18.7  & 39.5   & 32.9 \\
    BEVSegFormer~\cite{peng2023bevsegformer} & 800$\times$448 & Camera & ResNet-101       & 51.1 & 32.6  & 50.0   & 44.6  \\
    BEVFormer$^{\dag}$~\cite{li2022bevformer} & 1600$\times$900  & Camera & ResNet-50          & 53.0     & 36.6   & 54.1      & 47.9  \\
    BEVerse~\cite{zhang2022beverse} & 1408$\times$512 & Camera & Swin-Tiny  & 56.1 & 44.9    & 58.7   & 53.2  \\
    
    UniFusion~\cite{qin2022unifusion} & 1600$\times$900  & Camera & Swin-Tiny  & 58.6     & 43.3   & 59.0      & 53.6  \\
    
    \midrule
    HDMapNet-Fusion~\cite{li2022hdmapnet} & 352$\times$128  & Camera \& LiDAR & EfficientNet-B0 \& PointPillars  & 46.1     & 31.4  & 56.0      & 44.5  \\
    \midrule
    HDMapNet-LiDAR~\cite{li2022hdmapnet} & -  & LiDAR & PointPillars  & 26.7     & 17.3  & 44.6      & 29.5  \\
    LiDAR2Map  & -  & LiDAR & PointPillars &  60.4     &  45.5 & 66.4    & 57.4 \\
    LiDAR2Map & -  & LiDAR & VoxelNet  & \textbf{61.5}     & \textbf{46.3}  & \textbf{68.1}     & \textbf{58.6}  \\
    \bottomrule
    \end{tabular}}
\end{center}
\vspace{-4mm}
\end{table*}

\begin{table*}[ht]
\begin{center}
\caption{Performance comparison on the validation set of nuScenes with two commonly used settings for vehicle segmentation without masking invisible vehicles.
Setting 1 is with the $100\mathrm{m}\times50\mathrm{m}$ at $25$cm resolution. Setting 2 is with the $100\mathrm{m}\times100\mathrm{m}$ at $50$cm resolution.}

\label{tab:vehicle}
\scalebox{0.85}{
\begin{tabular}{l@{}c@{\ \ \ }c@{\ \ \ }c@{\ \ \ }c@{\ \ \ }c@{\ \ \ }c@{\ \ \ }c}
\toprule
Method  & Image Size & Modality & Backbone  & Setting 1 & Setting 2 & \#Params(M) & FPS   \\
\midrule
VED~\cite{lu2019monocular}  & 800$\times$600 & Camera & ResNet-50  & 8.8 & -  & -& -  \\
PON~\cite{roddick2020predicting} & 800$\times$600  & Camera & ResNet-50  & 24.7 & -  &  38 & 30 \\
VPN~\cite{pan2020cross}  & 800$\times$600 & Camera & ResNet-50  & 25.5 & -  & 18 & -  \\
STA~\cite{saha2021enabling} &  1280$\times$720 & Camera & ResNet-50  & 36.0 & - & - & -    \\
Lift-Splat~\cite{philion2020lift} & 352$\times$128  & Camera & EfficientNet-B0  & - & 32.1 & 14 & 25   \\
FIERY Static~\cite{hu2021fiery} & 448$\times$224  & Camera & EfficientNet-B4  & 37.7 & 35.8 & 7.4 & 8 \\
PolarBEV~\cite{liu2022vision} & 960$\times$448  & Camera & EfficientNet-B4  & 45.4 & 41.2 & 7.4 & 10 \\
SimpleBEV~\cite{harley2022simple} &  800$\times$448  & Camera & ResNet-101  & - & 47.4 & 37 & 7.3 \\
\midrule
TransFuseGrid~\cite{salazar2022transfusegrid} & 352$\times$128 & Camera \& LiDAR & EfficientNet-B0 \& PointPillars  & -  & 35.9 & -  & 18.4\\
\midrule
Pillar feature Net~\cite{salazar2022transfusegrid} & - & LiDAR & PointPillars  & - & 23.4 & - & - \\
LiDAR2Map & - & LiDAR & PointPillars  & \textbf{58.9} & \textbf{52.1} & 8.8 & \textbf{35}  \\

\bottomrule
\end{tabular}
}
\end{center}
\vspace{-6mm}
\end{table*}

\begin{figure*}
\centering
\includegraphics[width=0.9\linewidth]{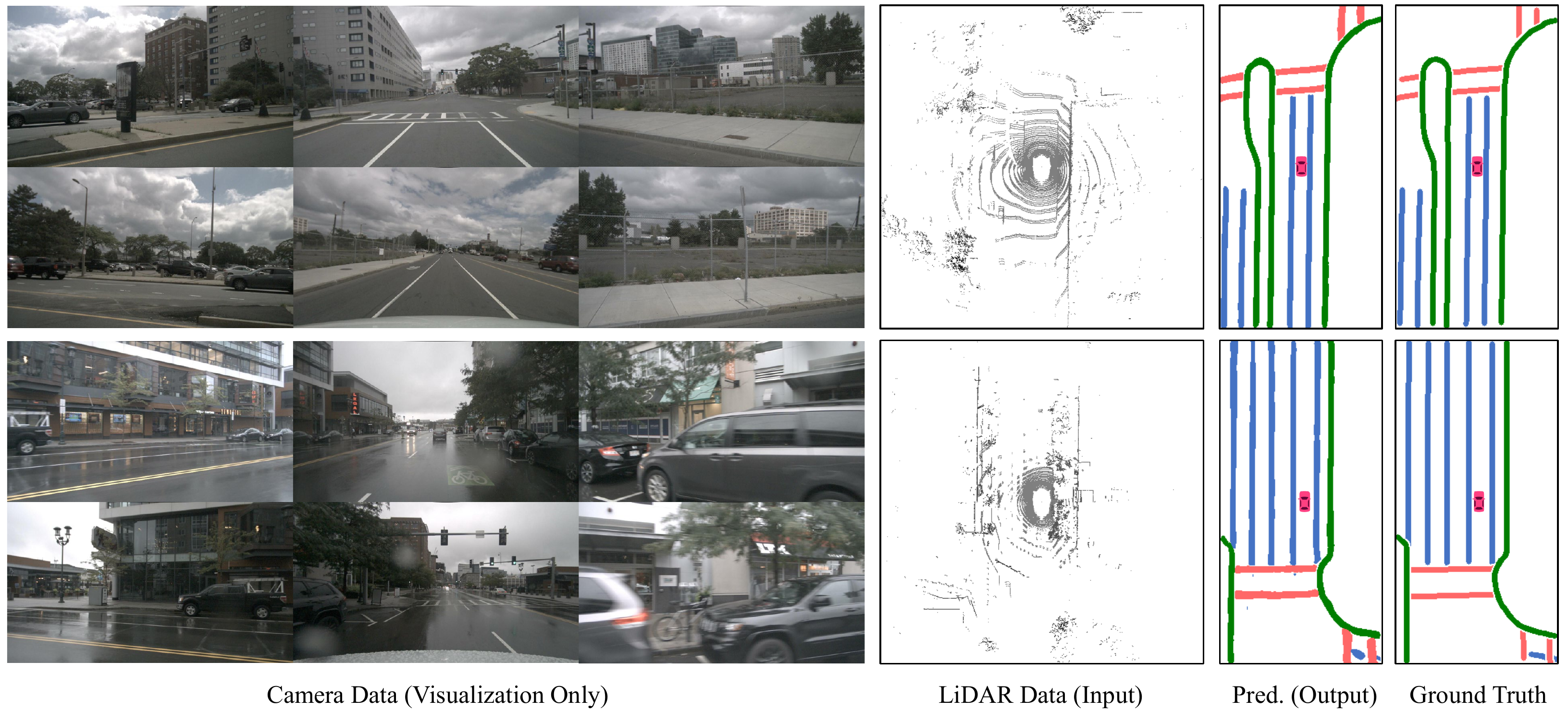}
    \caption{Visualization of LiDAR2Map on the validation set of nuScenes with cloudy and rainy condition scenes. The left shows the surrounding views from cameras with the six views, which are just used for visualization. The middle is the input LiDAR data for inference. The right is the predicted semantic map and the corresponding ground truth.}
    \label{fig:vis}
    \vspace{-4mm}
\end{figure*}

\subsection{Implementation Details}
\noindent \textbf{Dataset.}
To evaluate the efficacy on semantic map construction,
we conduct comprehensive experiments on nuScenes benchmark~\cite{caesar2020nuscenes} that is a general and authoritative dataset. It contains 1,000 driving scenes collected in Boston and Singapore. The vehicle used for data collection is equipped with a 32-beam LiDAR, five long range RADARs and six cameras. There are 700 and 150 complete scenes for training and validation, respectively.

\noindent \textbf{Evaluation.}
In this paper, we evaluate the performance on map and vehicle segmentation under different evaluation settings. For map segmentation, we adopt the same setting as HDMapNet~\cite{li2022hdmapnet}, which uses a $60\text{m} \times 30\text{m}$ area around the ego vehicle and samples a map at a $15\text{cm}$ resolution with three classes, including Divider (Div.), Ped Crossing (P. C.) and Boundary (Bound.). 
For vehicle segmentation, we utilize two commonly used settings proposed in PON~\cite{roddick2020predicting} and Lift-Splat~\cite{philion2020lift}. Setting 1 for vehicle segmentation employs a $100\text{m} \times 50\text{m}$ map around the ego vehicle and samples at a $25\text{cm}$ resolution. Setting 2 adopts a  $100\text{m} \times 100\text{m}$ map at $25\text{cm}$ resolution. 
The mean Intersection-over-Union (mIoU) is used for the performance evaluation.

\noindent \textbf{Training.}
For camera-branch, we choose Swin-Tiny~\cite{liu2021swin} pre-trained on ImageNet~\cite{russakovsky2015imagenet} as the image backbone.
For LiDAR-branch, PointPillars~\cite{lang2019pointpillars} and VoxelNet~\cite{zhou2018voxelnet} are used to extract the point cloud feature. We train the whole network with 30 epochs using Adam optimizer~\cite{kingma2014adam} having a weight decay of 1$e^{-7}$ on 4 NVIDIA Tesla V100 GPUs.  The learning rate is 2$e^{-3}$ for PointPillars and 1$e^{-4}$ for VoxelNet, which decreases with a factor of 10 at the 20th epoch.
The image size is set to  $352 \times 128$ for PointPillars and $704 \times 256$ for VoxelNet during training. 
More training details under different settings are given in our supplementary material.

\subsection{Main Results}
\noindent \textbf{Map Segmentation.}
For quantitative evaluation, we compare our method with the state-of-the-art camera-based models, including  BEVSegFormer~\cite{peng2023bevsegformer}, BEVFormer~\cite{li2022bevformer}, BEVerse~\cite{zhang2022beverse} and UniFusion~\cite{qin2022unifusion}, as shown in Tab.~\ref{tab:mapseg}. 
LiDAR2Map outperforms all the existing methods significantly and boosts the performance of the LiDAR-based models from 29.5\% mIoU to 57.4\% mIoU. Our model with PointPillars~\cite{lang2019pointpillars} outperforms the state-of-the-art camera-based methods by 3.8\% mIoU. With the stronger backbones like VoxelNet~\cite{zhou2018voxelnet}, LiDAR2Map even achieves a segmentation accuracy of 58.6\% mIoU. 
It is worthy of noting that LiDAR2Map achieves the promising results in the case of Boundary class. It indicates that the accurate height information from LiDAR is important for map segmentation. 
Furthermore, we visualize the results of LiDAR2Map in some typical driving scenarios including cloudy and rainy conditions as shown in Fig.~\ref{fig:vis}. 
More visualization results are included into the supplementary material.

\noindent \textbf{Vehicle Segmentation.}
Vehicle segmentation is one of the most important task among the moving elements in autonomous driving. 
In order to examine the scalability of our method, we evaluate LiDAR2Map under two different settings for vehicle segmentation. We only adopt PointPillars as the LiDAR backbone and report the inference speed of LiDAR2Map on single NVIDIA RTX 2080Ti GPU for a fair comparison. 
As shown in Tab.~\ref{tab:vehicle}, our method not only outperforms the state-of-the-art camera-based models by a large margin in accuracy, but also has the small model parameters with 35 FPS speed in inference. These promising results indicate the efficacy of our proposed LiDAR2Map approach and defend the strength of LiDAR on semantic map construction. 
We provide visual results on vehicle segmentation in the supplementary material.

\subsection{Ablation Studies}
\begin{figure}
\centering
\setlength{\abovecaptionskip}{1mm}
\includegraphics[width=0.85 \linewidth]{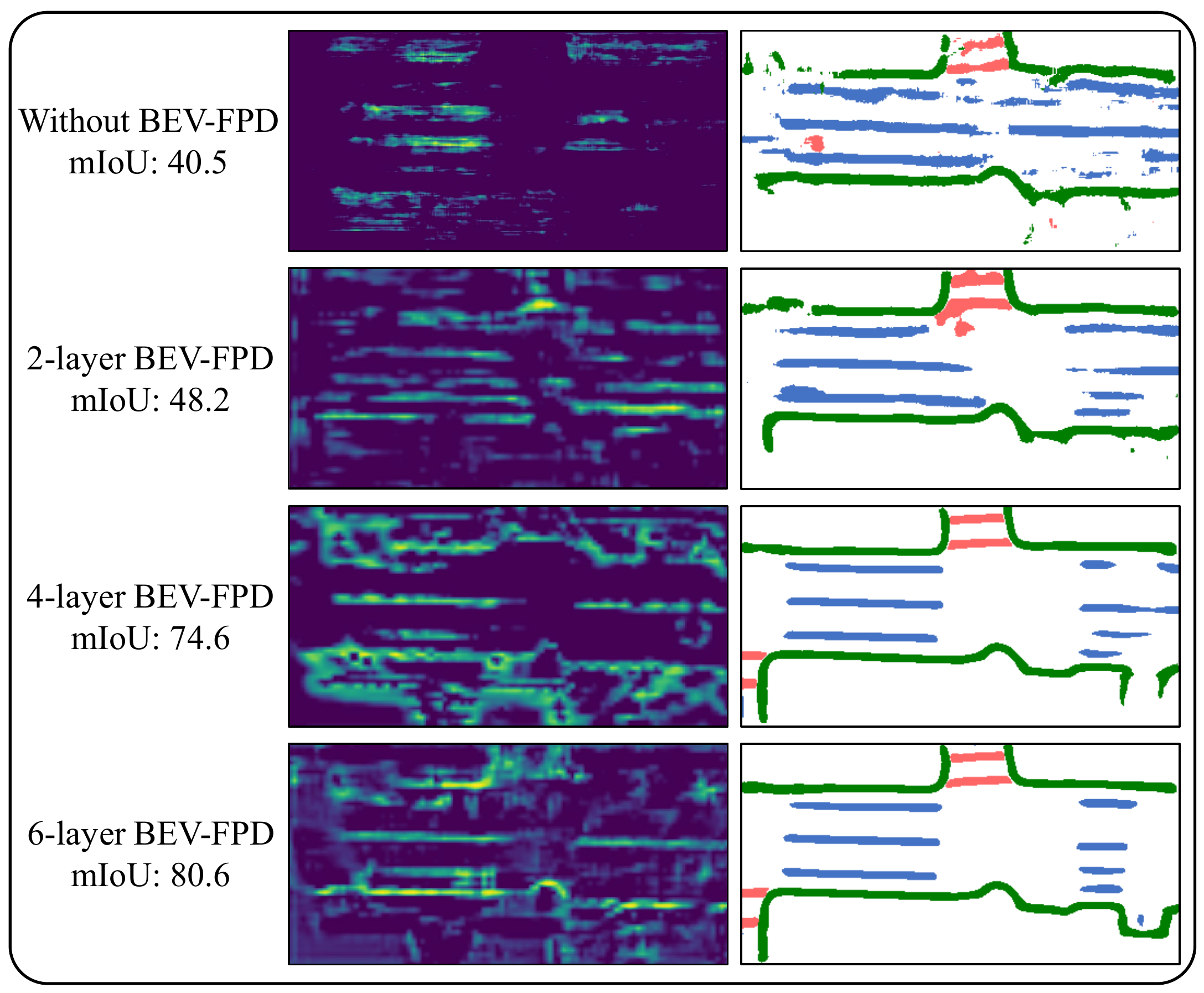}
    \caption{Visualization comparisons of LiDAR2Map with different BEV-FPDs and the corresponding semantic map predictions. The mIoU value means the evaluation score of the \textit{single frame}.
    Besides the baseline model without BEV-FPD, we provide the results of the second layer's output with 2-, 4- and 6-layer BEV-FPD based models, respectively. LiDAR2Map with 6-layer BEV-FPD obtains the best segmentation performance and its feature map has more accurate responses with less noises.}
    \label{fig:featmap}
    \vspace{-4mm}
\end{figure}

\noindent \textbf{BEV Feature Pyramid Decoder.}\label{ablation_bevdecoder}
In our experiments, we find that the layer number to obtain multi-scale features in the BEV-FPD has the substantial impact on the performance of LiDAR2Map for map segmentation. 
As shown in Tab.~\ref{tab:layer}, the results of Camera-LiDAR fusion model and LiDAR2Map using PointPillars have been greatly improved with the increasing number of layers. With the 2-layer model in BEV-FPD, our LiDAR2Map achieves 43.8\% mIoU. For the 4-layer model in BEV-FPD, a large performance improvement with +10.5\% mIoU is obtained, where LiDAR2Map achieves the comparable results against the recent camera-based methods like BEVerse~\cite{zhang2022beverse} and UniFusion~\cite{qin2022unifusion}. As the number of layers is increased to 6, the accuracy is boosted to 57.4\% mIoU and achieves the best performance.
We further visualize the feature maps to analyze our LiDAR2Map with different layer number in BEV-FPD.
As shown in Fig.~\ref{fig:featmap}, the model with 6-layer BEV-FPD holds the distinct response map in the region, where the target element appears with little noise for semantic map construction. Furthermore, Tab.~\ref{tab:layer} reports the performance of fusion model as the ``Teacher" in our LiDAR2Map. Notably, LiDAR2Map with 6-layer BEV-FPD as a ``Student" network has achieved the 98.8\% performance of fusion model with 2$\times$ faster inference speed.

\begin{table}[ht]
\centering
    \scalebox{0.95}{
		\begin{tabular}{c|ccc|c|c}
			\hline
			 Layer Num. & Div. & P. C. & Bound. & mIoU & FPS \\
			\hline
            \multirow{2}{*}{2}
            & 49.3 & 34.1 & 58.4 & 47.3 & 8.2  \\
            & \cellcolor{mygray}45.4 & \cellcolor{mygray} 30.5 & \cellcolor{mygray}55.6 & \cellcolor{mygray}43.8 & \cellcolor{mygray}23.3  \\
            \hline
   
            \multirow{2}{*}{4}
            & 56.9 & 45.1 & 64.0 & 55.3 & 7.2 \\
            & \cellcolor{mygray2}55.7 & \cellcolor{mygray2}43.9 & \cellcolor{mygray2}63.2 & \cellcolor{mygray2}54.3 & \cellcolor{mygray2}16.3 \\
            \hline
   
            \multirow{2}{*}{6}
            & 60.8 & 47.2 & 66.3 & 58.1 & 6.3 \\
            & \cellcolor{mygray3}60.4 & \cellcolor{mygray3}45.5 & \cellcolor{mygray3}66.4 & \cellcolor{mygray3}57.4 & \cellcolor{mygray3}12.6 \\
            \hline
	
			\hline
		\end{tabular}%
  }
	\caption{Accuracy and speed performance with  different layer number of BEV-FPD. At each row, the upper one is the results of the Camera-LiDAR fusion model (``Teacher"), and the lower one corresponds to the result of LiDAR2Map (``Student") in gray. }
     \label{tab:layer}
\end{table}

\noindent \textbf{Online Camera-to-LiDAR Distillation Scheme.}
To examine the effect of each module in the online Camera-to-LiDAR distillation, we conduct the ablation experiments on nuScenes, including map and vehicle segmentation. 
For vehicle segmentation, we adopt Setting 2 for performance evaluation. As shown in Tab.~\ref{tab:ablation}, our baseline model achieves 52.2\% mIoU on map segmentation by the design on 4-layer BEV-FPD.
The proposed Position-Guided Feature Fusion Module (PGF$^2$M) improves the baseline around 0.4\% mIoU and 1.5\% mIoU on map and vehicle, respectively. This demonstrates that multi-modality fusion is effective with both spatial features from LiDAR and semantic features from camera. Moreover, Feature-level Distillation (FD) and Logit-level Distillation (LD) achieve over 0.9/1.2\% mIoU and 1.1/0.7\% mIoU performance gains on map/vehicle segmentation, respectively. These encouraging results demonstrate that our proposed online distillation scheme can effectively improve the model accuracy.

\begin{table}[ht]
\centering
\setlength{\abovecaptionskip}{1mm}
    \setlength{\tabcolsep}{2.4mm}{
     \scalebox{0.97}{
		\begin{tabular}{c|ccc|c|c}
							\hline
                            Baseline & PGF$^2$M & FD & LD  & Map & Vehicle \\
							\hline
							\cmark &             &       &       & 52.2 & 49.1\\
							
							\cmark &  \cmark &       &       & 52.6& 50.6 \\
							\cmark &  \cmark & \cmark &       & 53.5 & 51.8 \\
                            \cmark &  \cmark &        & \cmark& 53.7 &51.3 \\
							
							\cellcolor{mygray2}\cmark &  \cellcolor{mygray2}\cmark & \cellcolor{mygray2}\cmark & \cellcolor{mygray2}\cmark & \cellcolor{mygray2}54.3 & \cellcolor{mygray2}52.1 \\
							\hline
						\end{tabular}}}
	\caption{The effectiveness of our online Camera-to-LiDAR distillation scheme with different settings on the nuScenes dataset.}
 \label{tab:ablation}
\vspace{-2mm}
\end{table}

\noindent \textbf{Comparison with Other Distillation Schemes.}
To further investigate the effectiveness of our online distillation scheme, we compare it with current knowledge distillation strategies. We have re-implemented these methods in the BEV feature space under the same setting to facilitate a fair comparison. Tab.~\ref{tab:kd} shows the comparison results. Among these methods, MonoDistill~\cite{chong2022monodistill} and MGD~\cite{yang2022masked} are feature-based distillation methods. Their results are even worse than the baseline model, which indicates the difficulty of the cross-modal knowledge distillation on high-dimensional BEV features. 
xMUDA~\cite{jaritz2020xmuda} and 2DPASS~\cite{yan20222dpass} are the logit-level distillation methods, which obtain better results over the baseline.
Our Camera-to-LiDAR distillation scheme provides a more effective way  compared against other distillation schemes and achieves the best performance.

\begin{table}[t]
\centering
\setlength{\abovecaptionskip}{1mm}
 \scalebox{0.95}{
		\begin{tabular}{l|ccc|c}
			\hline
			Method & Div. & P. C. & Bound. & mIoU\\
			\hline
			Baseline & 53.9 & 41.2  &  61.6 & 52.2  \\
			
            MonoDistill~\cite{chong2022monodistill} & 47.2 & 31.4 & 55.1 & 44.6  \\
            MGD~\cite{yang2022masked} & 52.0 & 38.7 & 59.6 & 50.1  \\
			xMUDA~\cite{jaritz2020xmuda} & 54.7 & 42.6 & 62.5 & 53.3  \\
			2DPASS~\cite{yan20222dpass}  & 55.3 & 43.0 & 62.4 & 53.6  \\
			\hline
			\cellcolor{mygray2}LiDAR2Map (Ours)   & \cellcolor{mygray2}55.7 & \cellcolor{mygray2}43.9 & \cellcolor{mygray2}63.2 & \cellcolor{mygray2}54.3  \\
							\hline
		\end{tabular}}
	\caption{Performance comparison with different knowledge distillation strategies on the nuScenes dataset.\vspace{-0.08cm}}
 \label{tab:kd}

\end{table}

\noindent \textbf{Different Number of Cameras.}
Tab.~\ref{tab:camera} reports the results to compare the performance with the camera branch using the different number of cameras.
The performance is not linearly related to the number of camera like those camera-based methods~\cite{zhou2022cross}. The LiDAR2Map model with two cameras of front and rear performs the best with 54.5\% mIoU while the models with all six cameras achieves 54.3\% mIoU. These results show that it is unnecessary to use so many cameras when the LiDAR is adopted.

\begin{table}[t]
\centering
\setlength{\abovecaptionskip}{1mm}
    \setlength{\tabcolsep}{3.4mm}{
     \scalebox{0.95}{
		\begin{tabular}{c|ccc|c}
			\hline
			Cam. Num. & Div. & P. C. & Bound. & mIoU \\
			\hline
			0  & 53.9     & 41.2  & 61.6      & 52.2 \\
   
			1  & 55.4 & 43.1 & 63.3 & 53.9 \\
            \cellcolor{mygray3}2  & \cellcolor{mygray3}56.3 & \cellcolor{mygray3}43.7 & \cellcolor{mygray3}63.4 & \cellcolor{mygray3}54.5 \\
            4  & 56.0 & 43.1 & 63.0 & 54.0 \\
            \cellcolor{mygray2}6  & \cellcolor{mygray2}55.7 & \cellcolor{mygray2}43.9 & \cellcolor{mygray2}63.2 & \cellcolor{mygray2}54.3 \\
			\hline
		\end{tabular}}}
	\caption{Performance comparison with different camera number during training on the nuScenes dataset. \vspace{-0.08cm}}
    \label{tab:camera}
\end{table}

\subsection{Scene-Level Semantic Map Construction}
\begin{figure}
\centering
\setlength{\abovecaptionskip}{1mm}
\includegraphics[width=0.85 \linewidth]{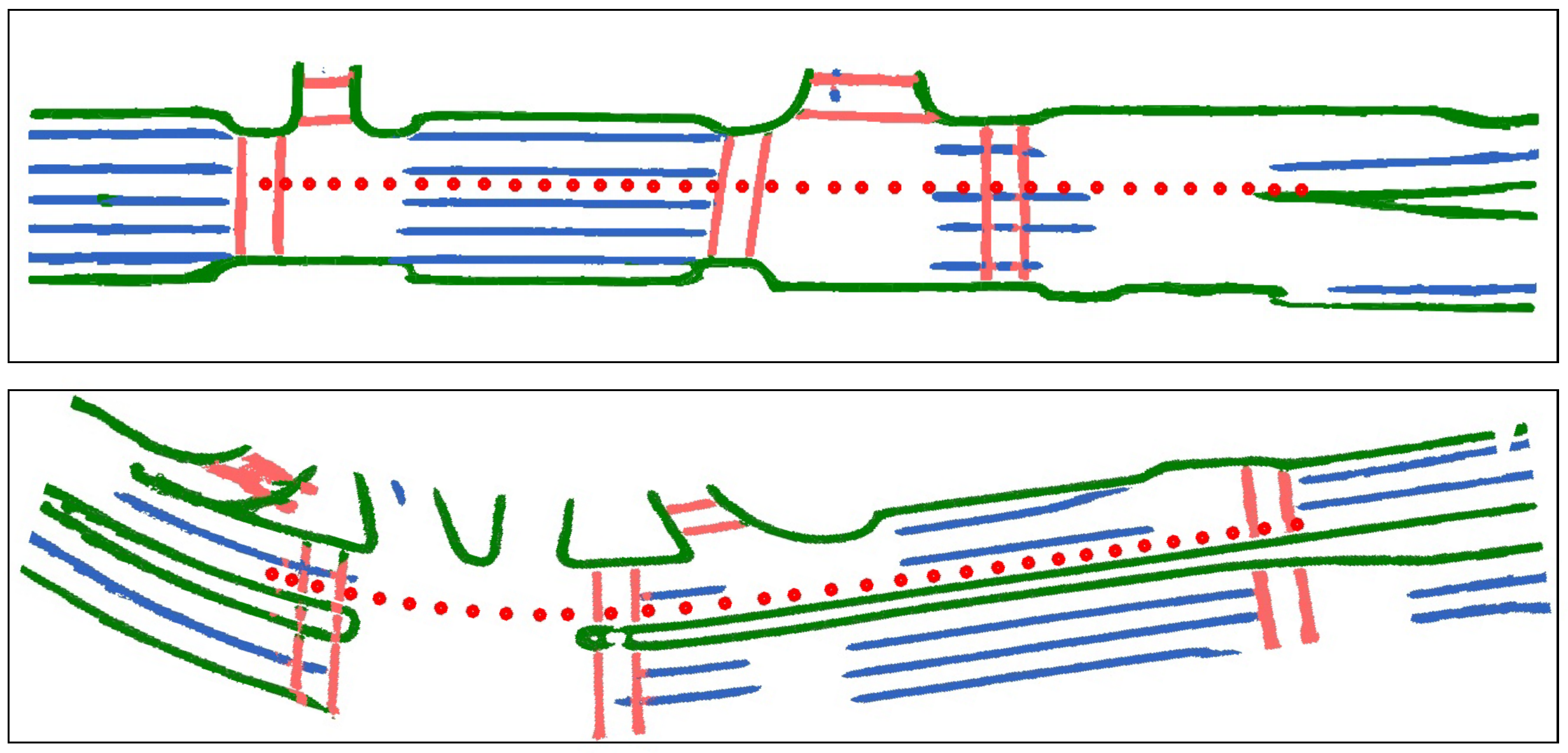}
    \caption{Scene-level semantic map obtained by accumulating 20s single frame maps with Bayesian filtering. The red dots indicate the trajectory of the ego vehicle.}
    \label{fig:scenemap}
    \vspace{-5mm}
\end{figure}

Semantic map construction in a single frame is limited for self-driving. It is necessary to fuse the keyframes in a whole scene for scene-level map construction. 
We construct the scene-level semantic map on nuScenes~\cite{caesar2020nuscenes}, which is a typical dataset collected in driving scenes. Each scene lasts for 20s, and around 40 keyframes are sampled at 2Hz.
We introduce a temporal accumulation method to build the scene-level semantic map. 
More precisely, the local semantic maps are warped to the global coordinate system with the extrinsic matrix. Then, the coincident regions are optimized by Bayesian filtering~\cite{thrun2002probabilistic, roddick2020predicting} to obtain a smooth global map.
The visual examples shown in Fig.~\ref{fig:scenemap} demonstrate that our LiDAR2Map approach is able to generate the consistent maps and provide more information for downstream tasks such as navigation and planning.

\section{Conclusion}

In this work, an efficient semantic map construction framework named LiDAR2Map, is presented with an effective BEV feature pyramid decoder and an online Camera-to-LiDAR distillation scheme. Unlike previous camera-based methods that have achieved excellent performance on this task, we mainly use LiDAR data and only extract image features as auxiliary network during training. The designed distillation strategy can make the LiDAR-based network well benefit from the semantic features of the camera image. Eventually, our method achieves the state-of-the-art performance on semantic map construction including map and vehicle segmentation under several competitive settings. The distillation scheme in LiDAR2Map is a general and flexible cross-modal distillation method. In the future, we will explore its application in more BEV perception tasks such as 3D object detection and motion prediction.

\section*{Acknowledgments}
This work is supported by National Natural Science Foundation of China under Grants (61831015). It is also supported by Information Technology Center and State Key Lab of CAD\&CG, Zhejiang University.

{\small
\bibliographystyle{ieee_fullname}
\bibliography{egbib}
}


\clearpage
\appendix
\section*{Appendix}

\setcounter{figure}{0}
\setcounter{table}{0}
\renewcommand{\thefigure}{A\arabic{figure}}
\renewcommand{\thetable}{A\arabic{table}}

\section{Training for Vehicle Segmentation}
\label{sec:imple}
The training details for vehicle segmentation in Setting 1 and Setting 2 are slightly different from map segmentation.
Also, we adopt Swin-Tiny~\cite{liu2021swin} and PointPillars~\cite{lang2019pointpillars} as the feature extractors for image and LiDAR point cloud, respectively. The BEV feature pyramid decoder (BEV-FPD) uses a three-layer model with a trade-off between the accuracy and inference speed. 
We train the whole network for 15 epochs with 2 NVIDIA RTX 2080Ti GPUs.  
The learning rate is 1.5$e^{-3}$, which decreases by a factor of 10 at the 10th epoch.
The image size is set to $352 \times 128$ during training.

\section{Additional Results}
\label{sec:add}
\subsection{Map Segmentation}

\begin{figure*}
\centering
\includegraphics[width=0.83\linewidth]{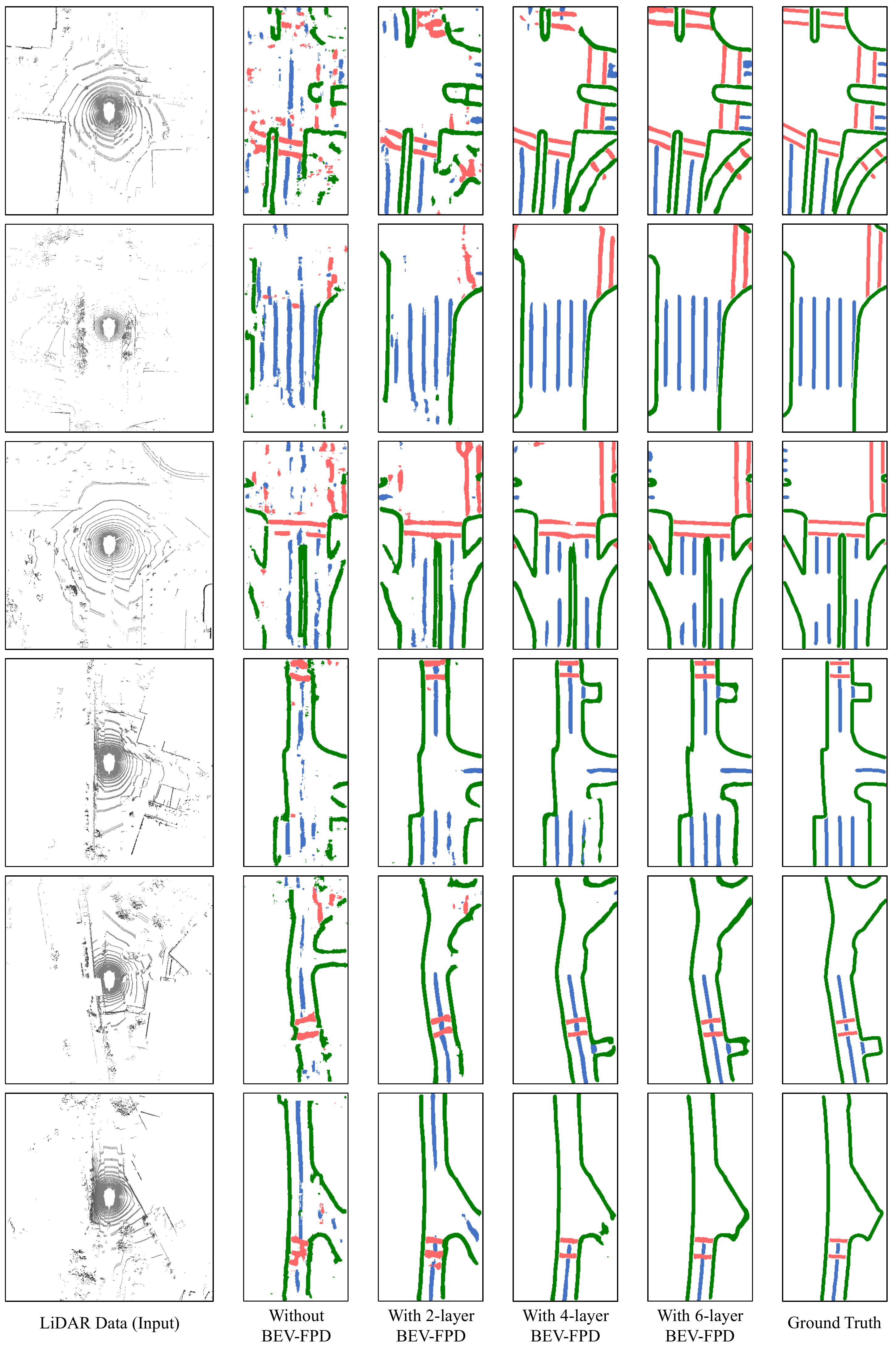}
    \caption{Additional visualization comparisons of LiDAR2Map with different BEV-FPDs on the nuScenes dataset.}
    \label{fig:vis_fpd}
\end{figure*}

\noindent \textbf{More Visual Results for BEV-FPD.} 
We provide more visual results from the output of LiDAR2Map with different BEV-FPDs. 
In Fig.~\ref{fig:vis_fpd}, the predicted semantic maps
are gradually refined and become more accurate 
with the deepening of the number of layers, which further indicates the effectiveness of BEV-FPD on promoting our LiDAR2Map.

\begin{figure*}
\centering
\includegraphics[width=0.625\linewidth]{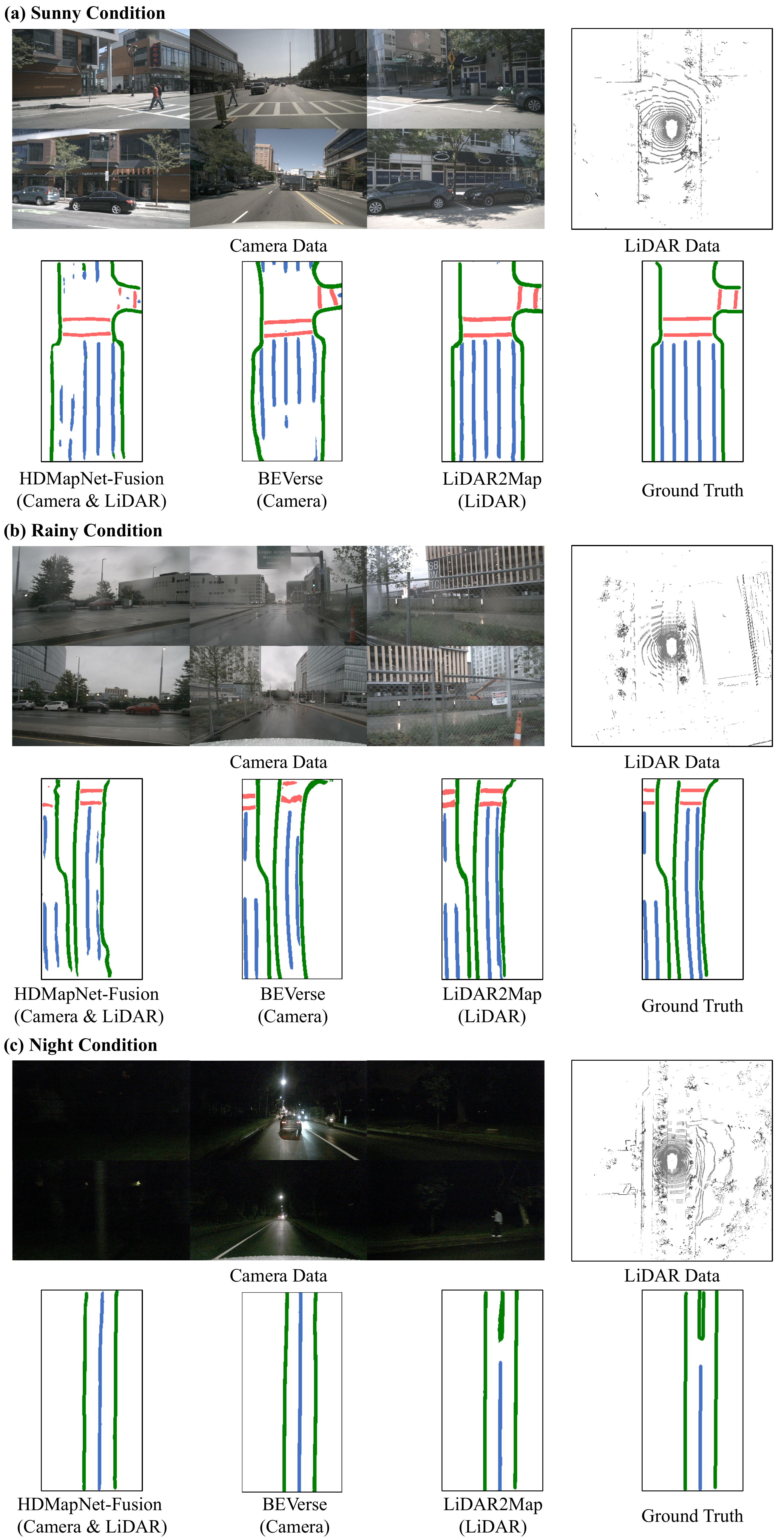}
    \caption{Qualitative results under various conditions. We compare our LiDAR2Map with other advanced approaches, including HDMapNet-Fusion~\cite{li2022hdmapnet} and BEVerse~\cite{zhang2022beverse}.}
    \label{fig:vis_comp}
\end{figure*}

\begin{figure*}
\centering
\includegraphics[width=0.94\linewidth]{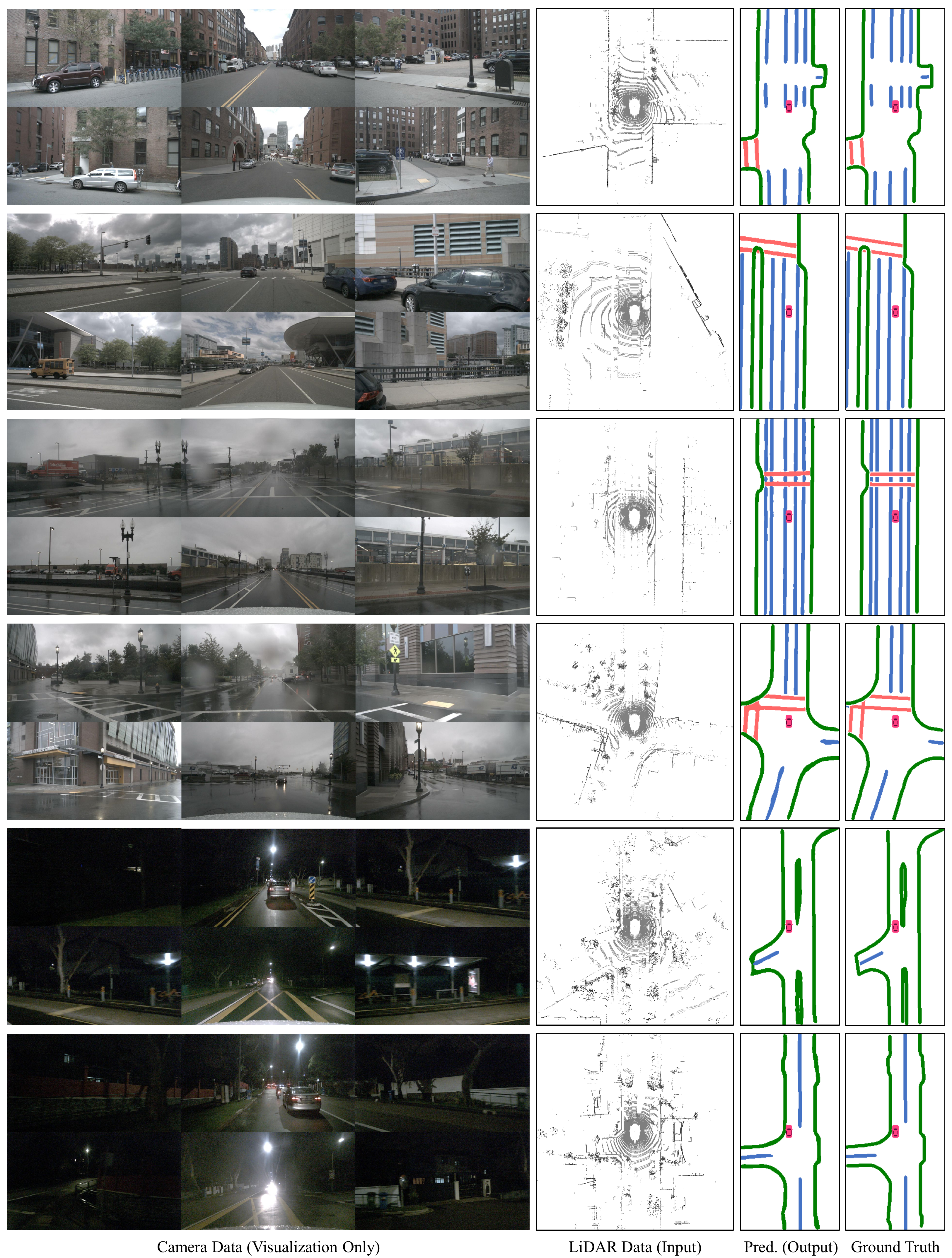}
    \caption{Additional visualization on map segmentation of LiDAR2Map with cloudy/rainy and day/night condition scenes.}
    \label{fig:vis_supp}
\end{figure*}

\noindent \textbf{Comparison Under Different Weather and Light Conditions.} 
As illustrated in Tab.~\ref{tab:weather}, we compare LiDAR2Map with the state-of-the-art methods including HDMapNet-Fusion~\cite{li2022hdmapnet} and BEVerse~\cite{zhang2022beverse} in different conditions. We employ PointPillars~\cite{lang2019pointpillars} as LiDAR backbone and 6-layer BEV-FPD for LiDAR2Map. Our method achieves the stable segmentation accuracy and outperforms other methods under different weather and light conditions. Fig.~\ref{fig:vis_comp} provides the qualitative comparison in several typical scenarios. 
LiDAR2Map presents the superior capability in sunny, rainy and nighttime compared to  HDMapNet-Fusion~\cite{li2022hdmapnet} and BEVerse~\cite{zhang2022beverse}.
Fig.~\ref{fig:vis_supp} further reports more map predictions of our LiDAR2Map.

\begin{table}[ht]
\centering
    \scalebox{0.8}{
		\begin{tabular}{c|c|ccc}
			\hline
			Method & Modality & Rainy & Night  & All \\
			\hline
			HDMapNet-Fusion~\cite{li2022hdmapnet} & Camera \& LiDAR &  38.7    &  39.3    & 44.5 \\
            BEVerse$^{*}$~\cite{zhang2022beverse}  & Camera &  48.8    &  44.4   & 51.7 \\
   
			\cellcolor{mygray3}LiDAR2Map (Ours)  & \cellcolor{mygray3}LiDAR & \cellcolor{mygray3}49.6  & \cellcolor{mygray3}49.2 & \cellcolor{mygray3}57.4 \\
			\hline
		\end{tabular}%
  }
	\caption{Map segmentation results under different weather and light conditions on the nuScenes dataset. ``$*$'' means the results are obtained from its official public model. }
    \label{tab:weather}
\end{table}

\subsection{Vehicle Segmentation}
\begin{figure*}
\centering
\includegraphics[width=1.0\linewidth]{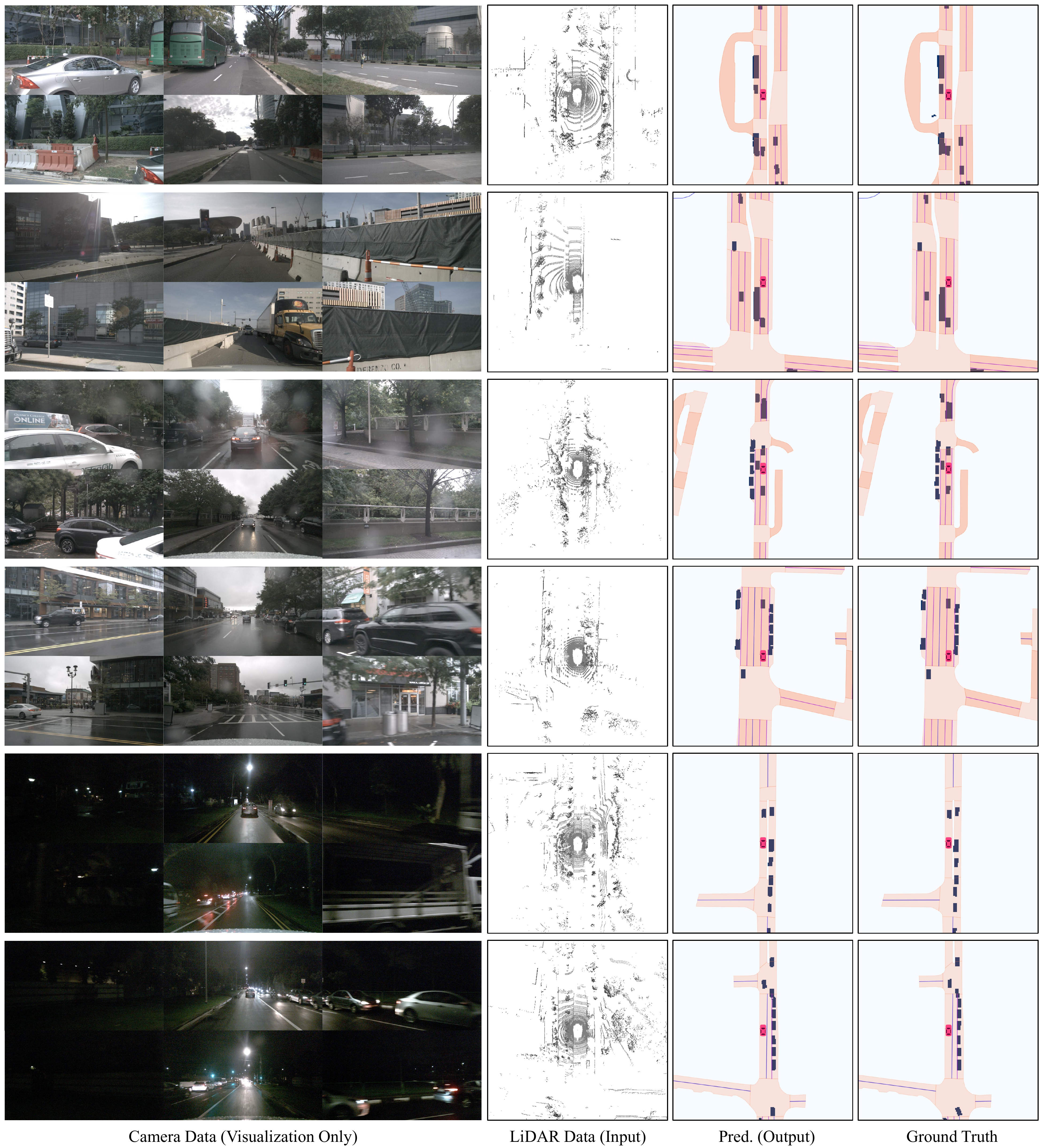}
    \caption{Visual vehicle segmentation of LiDAR2Map with cloudy/rainy and day/night condition scenes.}
    \label{fig:vis_veh}
\end{figure*}

For vehicle segmentation, we provide the qualitative results on the nuScenes dataset with Setting 2 in Fig.~\ref{fig:vis_veh}.
It obviously indicates that our method obtains the accurate vehicle predictions in different scenes.

\section{Limitations and Future Work}
The online Camera-to-LiDAR distillation scheme in our method incurs a certain amount of computation during the training, which increases the overall training time.
Besides, the semantic map construction task relies on high-definition map annotations for the network training, which are only available in few datasets~\cite{caesar2020nuscenes}. 
This limits the application of semantic map to more general autonomous driving scenarios.
In the future, we will try to speed up the training process and explore the potential of LiDAR2Map with weakly-supervised forms, such as open street map~\cite{haklay2008openstreetmap}.

\end{document}